\documentclass[acmsmall,screen]{acmart}
\AtBeginDocument{%
  }

\setcopyright{acmlicensed}
\copyrightyear{2018}
\acmYear{2018}
\acmDOI{XXXXXXX.XXXXXXX}

\acmJournal{JACM}
\acmVolume{37}
\acmNumber{4}
\acmArticle{111}
\acmMonth{8}

\usepackage{svg}
\usepackage{float}
\usepackage{natbib}
\usepackage{graphicx}
\usepackage{wrapfig}
\usepackage{subcaption}
\begin{document}

\title{S4ConvD: Adaptive Scaling and Frequency Adjustment for Energy-Efficient Sensor Networks in Smart Buildings}


\author{Melanie Schaller}
\orcid{0000-0002-5708-4394}
\email{schaller@tnt.uni-hannover.de}
\author{Bodo Rosenhahn}
\orcid{0000-0003-3861-1424}
\email{rosenhahn@tnt.uni-hannover.de}

\affiliation{%
  \institution{Institute for Information Processing (tnt), Leibniz University Hannover}
  \city{Hannover}
  \state{Lower Saxony}
  \country{Germany}
}

\affiliation{%
  \institution{L3S Research Center}
  \city{Hannover}
  \state{Lower Saxony}
  \country{Germany}
}

% \affiliation{%

% \affiliation{%

\renewcommand{\shortauthors}{Schaller et al.}

\begin{abstract}
Predicting energy consumption in smart buildings is challenging due to dependencies in sensor data and the variability of environmental conditions. We introduce S4ConvD, a novel convolutional variant of Deep State Space Models (Deep-SSMs), that minimizes reliance on extensive preprocessing steps. S4ConvD is designed to optimize runtime in resource-constrained environments. By implementing adaptive scaling and frequency adjustments, this model shows to capture complex temporal patterns in building energy dynamics. Experiments on the ASHRAE Great Energy Predictor III dataset reveal that S4ConvD outperforms current benchmarks. Additionally, S4ConvD benefits from significant improvements in GPU runtime through the use of Block Tiling optimization techniques. Thus, S4ConvD has the potential for practical deployment in real-time energy modeling. Furthermore, the complete codebase and dataset are accessible on GitHub, fostering open-source contributions and facilitating further research. Our method also promotes resource-efficient model execution, enhancing both energy forecasting and the potential integration of renewable energy sources into smart grid systems.
\end{abstract}




\received{28 February 2025}
\received[revised]{12 March 2025}
\received[accepted]{5 June 2025}


\begin{teaserfigure}
    \centering
    \includegraphics[width=0.85\textwidth]{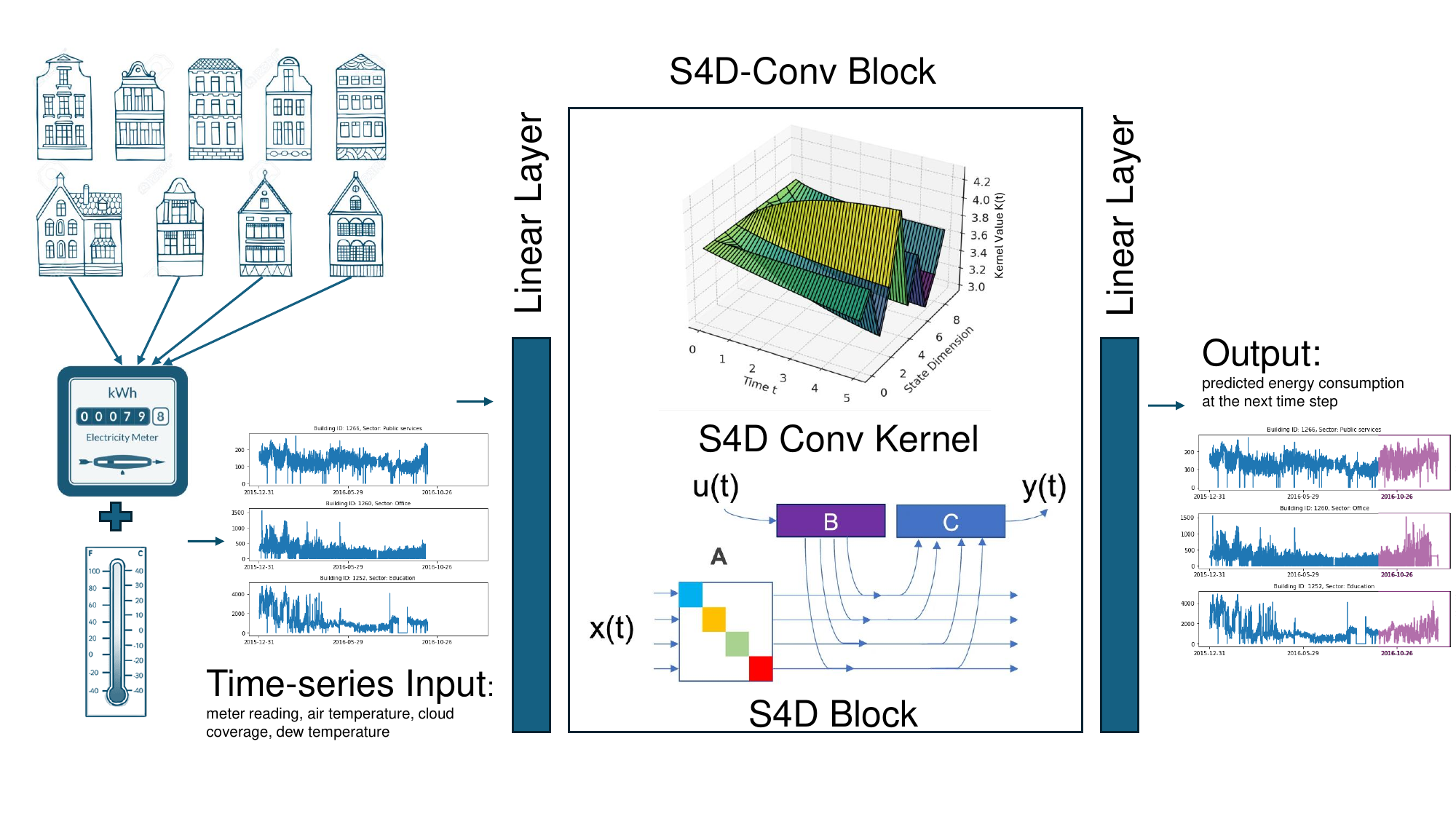} 
    \Description{Visualization of the processing pipeline: The multivariate time-series with meter-readings and temperature features inputs, and the S4ConvD layer's kernel visualization.}
    \caption{
      Processing pipeline: Meter-readings and temperature inputs are encoded linearly. The S4ConvD layer's kernel visualization shows:
  \textbf{x-Axis}: Temporal development.
  \textbf{y-Axis}: State-space components.
  \textbf{z-Axis}: Kernel values. 
  Concludes with a Decoder predicting energy consumption (best viewed in color and zoomed in.)
    }
    \label{fig:Fig1}
\end{teaserfigure}

\maketitle

\section{Introduction}
% \begin{figure}[ht]
%     \centering
%     \includegraphics[width=1.0\textwidth]{Fig1.pdf} 
%     \Description{Visualization of the processing pipeline.}
%     \caption{
%       Visualization of the processing pipeline: The multivariate time-series, consisting of meter-readings and temperature features, is used as input. A linear layer functions as the Encoder. Our proposed S4ConvD layer is depicted with its kernel visualized over time. 
%       \textbf{x-Axis (Time t)}: Illustrates temporal development of the kernel. 
%       \textbf{y-Axis (State Dimension)}: Distinguishes various state-space components. 
%       \textbf{z-Axis (Kernel Value $K(t)$)}: Represents kernel values influenced by $e^{tA}B$, showing exponential changes with damping based on eigenvalues of $A$. Additional nonlinear effects may occur due to the scaling function $\sigma(\cdot)$. The pipeline proceeds with a linear layer as the Decoder, culminating in the output of predicted energy consumption.
%       (Best viewed in color and zoomed in for detail.)
%     }
%     \label{fig:Fig1}
% \end{figure}

The field of building energy consumption prediction with sensor networks~\cite{kazmi2014review, erickson2014occupancy} has made significant advancements over the past decades with the improvement of data-driven approaches~\cite{amasyali2018review, olu2022building, li2017building} and smart metering infrastructure~\cite{yildiz2017recent}. These have enabled the collection of vast amounts of data and the generation of valuable insights through machine learning~\cite{siryani2017machine, elsisi2021reliable}. Initially, research in the 1990s utilized statistical models and machine learning techniques~\cite{miller, ashrae-energy-prediction} such as artificial neural networks (ANN)~\cite{ANN} and support vector machines (SVM)~\cite{SVM}, which have later evolved into more complex ensemble methods that reduce errors through model diversity. Despite numerous studies, comparing different prediction methods is challenging due to their customization for specific contexts~\cite{ashrae-energy-prediction}. While numerous building energy prediction methods have been devised over the past thirty years~\cite{ASLAM2021110992}, there remains no widespread agreement on the most effective methods for different types of buildings. Moreover, many of the developed techniques are not readily accessible to the broader research community. It has been observed, that a majority with 33 out of 42 of prediction studies using measured data implemented and tested the models on a single building~\cite{miller}.

To address this issue, the ASHRAE Great Energy Predictor III (GEPIII) Competition~\cite{miller, miller2022gradient, ashrae-energy-prediction}, held in 2019, sought to enhance building energy consumption forecasting using machine learning models for different types of buildings. This competition series, which began in 1993, has served as a cornerstone for crowdsourced benchmarking of time-series data in the building industry and is therefore still used. The competition aimed to identify the best methods for hourly energy prediction in commercial buildings and disseminate the accumulated modeling knowledge to the wider academic and practitioner communities~\cite{ashrae-energy-prediction}.

In recent years, integrating physical process representations into neural networks has enhanced the modeling of dynamic systems by capturing temporal dependencies in real-world data~\cite{schaller2024modeconv, schaller2023liquor}. State Space Models (SSMs) have emerged as a powerful alternative to Transformer architectures~\cite{gu2021efficiently, gu2023mamba, gupta2022diagonal, smith2023simplified, wang2024graph}, offering efficient handling of long-range dependencies while maintaining lower computational complexity. As first Deep SSM, the Structured State Space Models (S4)~\cite{gu2021efficiently} and its derivative, S4D~\cite{gupta2022diagonal} have been introduced. The S4 framework introduced the handling of sequential data by utilizing the structured state space representation's using HiPPO (High-order Polynomial Projection Operator) matrices' inherent ability to capture long-range dependencies~\cite{gu2021efficiently}. Building upon S4, S4D further refines this approach by introducing diagonal matrices that enhance the model's numerical robustness~\cite{gupta2022diagonal}. This foundational work in S4 and S4D sets the basis for our proposed S4ConvD model, a novel extension of S4D for energy consumption prediction. Unlike the standard S4D convolution with its Cauchy Kernel approach, S4ConvD makes use of a convolutional kernel, to integrate adaptive state-space modeling and thus to dynamically adjust to changing energy patterns, enhancing both accuracy and runtime.

The main contributions of this research are:
\begin{enumerate}
\item \textbf{Adaptive State-Space Convolution:} S4ConvD extends the S4D framework by introducing a dynamically parameterized state matrix and an adaptive input transformation. 
\item \textbf{Efficient Frequency Sensitivity:} Through optimized handling of frequency components, S4ConvD selectively emphasizes relevant temporal patterns, improving performance over static convolutional methods without excessive computational overhead.
\item \textbf{Benchmark Validation:} Evaluations on the ASHRAE Great Energy Predictor III dataset show that S4ConvD outperforms competing models, demonstrating superior generalization across diverse building types.
\item \textbf{CUDA Memory Optimization:} We demonstrate that the CUDA optimization technique Block Tiling, specifically tailored for the S4ConvD methodology on modern GPU architectures, can improve runtime by 36 \%. 
\item \textbf{Reproducibility and Open Science:} To encourage further research and transparency, we provide the full code and dataset on GitHub\footnote{https://github.com/MilanShao/S4ConvD}.
\end{enumerate}

\section{Related Work}
We first list up the observations of the Great Energy Predictor III dataset challenge~\cite{gu2021efficiently, gu2023mamba, gupta2022diagonal, smith2023simplified, wang2024graph}, that were made by the winning teams. Then we summarize the methods used by the three winning teams of the challenge as benchmarking results. As last point we introduce the related work on deep SSMs.
One of the key differences between the top-performing approaches and approaches with worse performance was the used preprocessing methods to filter the data before modeling and the corrections they applied after prediction and before submission e.g for removing anomalous behavior from the dataset as well as various methods to apply weightings and the creation of complex ensembling frameworks~\cite{ashrae-energy-prediction}.

The first-place implemented comprehensive data preprocessing to remove anomalies, impute missing values, correct time zones and employed feature engineering to extract features, which included raw data, categorical interactions, temporal attributes, various weather features, and target encoding. An ensemble of CatBoost~\cite{dorogush2018catboost}, LightGBM~\cite{ke2017lightgbm}, and Multi-Layer Perceptrons (MLP)~\cite{gardner1998artificial} was used. The final predictions were obtained by combining individual model outputs using a generalized weighted mean approach~\cite{ashrae-energy-prediction}.

The second-place approach involved manually removing outliers through visual inspection and filtering of each building's data. For feature selection, simple statistical and temporal features from weather and building metadata were computed, without relying on complex lag features. XGBoost~\cite{chen2015xgboost}, LightGBM~\cite{ke2017lightgbm}, CatBoost~\cite{dorogush2018catboost} and Feed-forward Neural Networks (FFNN)~\cite{svozil1997introduction}, specifically used for the electrical meter~\cite{yildiz2017recent}, were used as ensemble~\cite{ashrae-energy-prediction}.

The third-place incorporated a log transformation on the target variable. The feature engineering involved weather-related features like heat, windchill, and lagged weather features, along with temporal data and building metadata. Models including CatBoost~\cite{dorogush2018catboost}, neural networks~\cite{ANN} and LightGBM~\cite{ke2017lightgbm} were trained. The final predictions were achieved by ensembling~\cite{dietterich2000ensemble} these models using a weighted average method, with weights derived from publicly available datasets~\cite{ashrae-energy-prediction}.

In building energy prediction, issues of generalizability and scalability remain significant challenges~\cite{miller}. Traditional machine learning models often struggle to adapt across various building types and contexts. This necessitates exploring deep learning approaches, which offer the potential to capture intricate patterns and temporal behaviors. Additionally, integrating data-driven models with physics-based approaches can provide a more holistic understanding of energy consumption dynamics~\cite{miller}. Motivated by these challenges and opportunities, the development of S4ConvD aims to bridge these gaps. 

While Transformer-based architectures have achieved remarkable success in sequence modeling~\cite{vaswani2023attentionneed}, their quadratic complexity~\cite{katharopoulos2020transformers} limits scalability. Recent advancements in State Space Models (SSMs)~\cite{gu2021efficiently, gu2023mamba, gupta2022diagonal, smith2023simplified, wang2024graph} provide an efficient alternative by utilizing structured state transitions to capture temporal dynamics with lower computational overhead. SSMs originally derive from control theory~\cite{hamilton1994state}. In control engineering and system identification, state-space representation models a physical system through input, output, and state variables linked by first-order differential or difference equations. State variables evolve over time based on their current values and externally applied inputs. Output variables depend on the state variables, and potentially the input variables~\cite{brogan1974modern, Harvey_1990}.  However, existing SSM approaches often rely on fixed parameterization, limiting adaptability to rapidly changing data distributions. S4ConvD addresses this gap by introducing an adaptive state-space convolution, dynamically parameterized state matrices, enabling enhanced frequency sensitivity. Our results demonstrate that these refinements lead to superior performance in energy consumption forecasting, particularly in urban-scale datasets such as the ASHRAE Great Energy Predictor III dataset.

\section{Dataset}

\begin{wrapfigure}{R}{0.5\textwidth}
    \centering
    \includegraphics[width=0.48\textwidth]{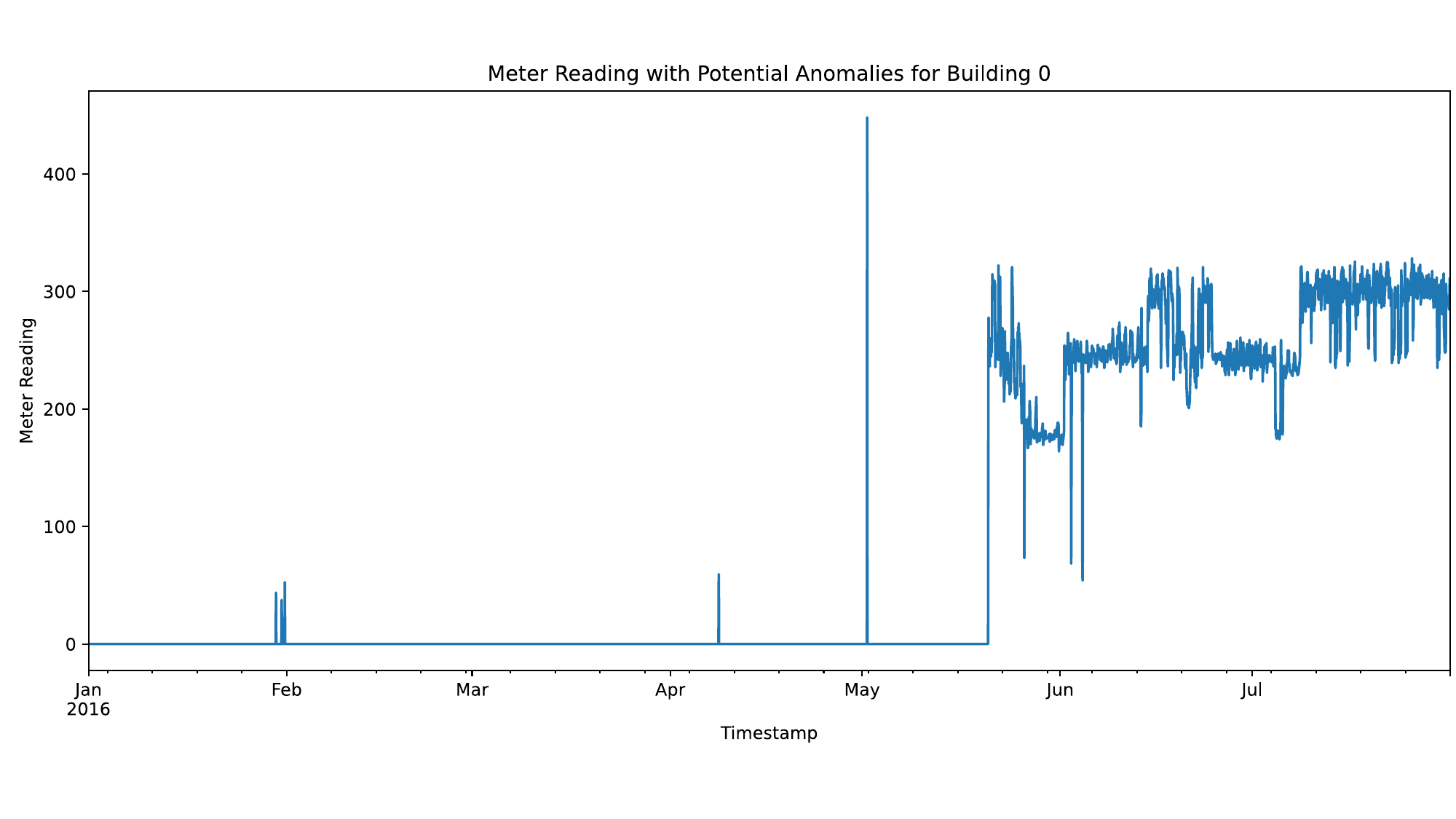}
    \Description{Example of an anomaly, highlighting unexpected patterns in the dataset.}
    \caption{Example of an anomaly in the dataset.}
    \label{fig:anomaly_example}
\end{wrapfigure}

Due to limitations in scalability observed in many current approaches, which often cater to specific or limited datasets~\cite{miller}, the Great Energy Predictor III dataset~\cite{miller, miller2022gradient, ashrae-energy-prediction} is used. This dataset is compiled from approximately 61,910,200 energy measurements acquired from diverse building types across 16 geographical locations worldwide. ASHRAE also hosted the Great Energy Predictor III (GEPIII) machine learning competition on the Kaggle platform 2019. This dataset captures hourly energy consumption metrics for 2,380 energy meters situated within 1,448 buildings. It contains various types of anomalies, such as abrupt changes in meter readings or outlier temperature values, which could be due to sensor malfunctions, data entry errors, or unusual operational days. See, for instance, Fig. 2, which depicts an example of such measurement anomalies where there is an unexpected spike in energy consumption not aligned with typical patterns and external conditions.
   
These hourly measurements come from different building types. The t-SNE (t-Distributed Stochastic Neighbor Embedding) distribution~\cite{wattenberg2016use} in Fig. 3 visualizes high-dimensional building data by reducing its dimensionality to two dimensions while preserving the local structure of the data. Each point in the plot represents a building, and its color indicates its primary use category. The legend provides a mapping between colors and categories, including lodging/residential, entertainment/public assembly, office, public services, education, parking, food sales and service, retail, warehouse/storage, other, healthcare, utility, technology/science, manufacturing/industrial, and religious worship. The distribution of points suggests that these building categories don't form dense clusters. The lack of dense clustering for certain categories could suggest variations in building design and operational efficiency which are not strictly bound by building type similarity. 

To support analysis, coincident weather data was provided for each site. The dataset underwent minimal cleaning and processing to reflect real-world conditions accurately. It covers four different types of energy consumption: electricity, chilled water, steam, and hot water. The modeling process involved examining historical usage patterns alongside corresponding weather data. 

Fig. 4 contains multiple time-series, each representing the mean meter readings of different buildings over time. The x-axis in all plots corresponds to the time period from March 2016 to July 2016, while the y-axis represents the mean meter reading. Each plot is labeled with a specific \textit{building\_id}, indicating the building for which the data is displayed. 

In each graph, two different aggregations are shown: the blue lines represent the readings recorded on an hourly basis, capturing short-term fluctuations, whereas the orange lines depict daily averages, providing a smoothed representation of the overall trend. Most buildings exhibit an increasing trend in energy consumption over time, though the degree of variability differs. Some buildings, such as those with \textit{building\_id: 60}, display sudden spikes, suggesting anomalies or possible data recording issues. In contrast, buildings such as \textit{building\_id: 76} show high fluctuations, but with a clear increasing trend. The daily aggregation follows the overall shape of the hourly data while reducing noise, making it easier to observe long-term trends.  

\begin{figure}[htbp]
    \centering
    \includegraphics[width=1.0\textwidth]{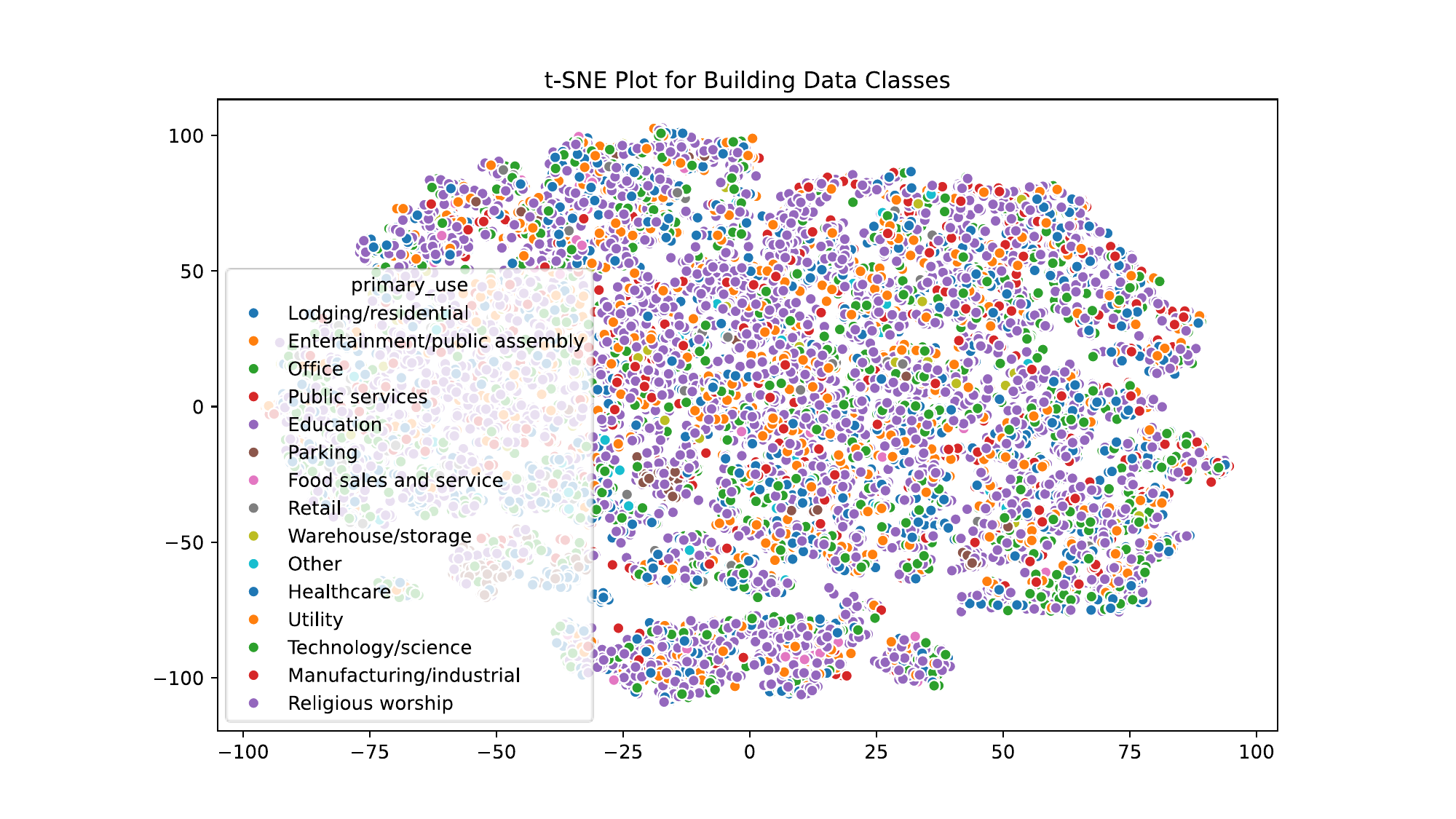}
    \Description{t-SNE.}
    \caption{t-SNE (t-Distributed Stochastic Neighbor Embedding) plot of distributions along building types.}
    \label{fig:t_SNE}
\end{figure}

\subsection{Datasplit}

The dataset was segmented into three disjunct separate sets. The dataset is divided into a training set consisting of 11,637,272 rows, a validation set comprising 3,415,636 rows, and a test set containing 5,108,845 rows. This should prevent from data leakage like it is caused in cross validation settings~\cite{schaller2025automlmulticlassanomalycompensation}.

\section{Evaluation Metric}
To evaluate the submissions in the ASHRAE Great Energy Predictor III competition~\cite{ashrae-energy-prediction, miller}, the Root Mean Squared Logarithmic Error (RMSLE)~\cite{hodson2022root} was used as the primary metric. RMSLE was chosen due to its capability to mitigate the disproportionate influence of meters with larger consumption values~\cite{gilroy1990mean}, thereby offering a balanced scoring approach compared to the conventional Root Mean Square Error (RMSE)~\cite{chai2014root}. This choice was particularly relevant since significant variations in meter readings could skew results if not addressed properly. Given the nonprofit nature of the competition, the team opted for this established metric over a custom-designed one.

\begin{figure}[H]
    \centering
    \includegraphics[width=1.0\textwidth]{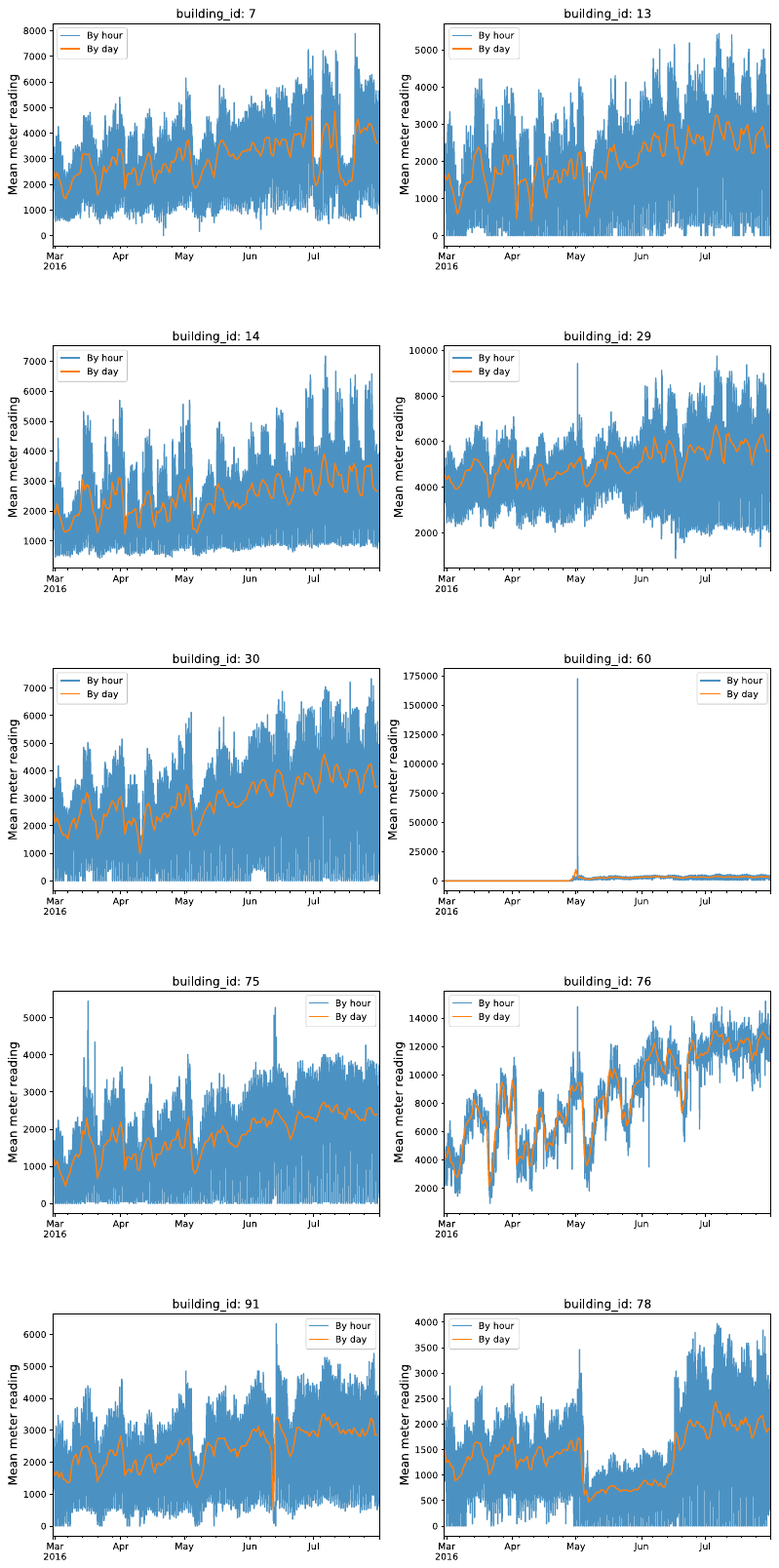} 
    \caption{Energy meter-readings for different exemplary buildings with mean values per hour and per day sorted by building-ID.}
    \label{fig:meterbuildingid}
\end{figure}

The RMSLE is defined by the equation~\cite{hodson2022root}:

\begin{equation}
\text{RMSLE} = \sqrt{\frac{1}{n} \sum_{i=1}^{n} \left(\log(\text{pred}_i + 1) - \log(\text{actual}_i + 1)\right)^2}
\end{equation}

where \( \text{pred}_i \) represents the predicted value, \( \text{actual}_i \) the actual value, and \( n \) is the number of observations~\cite{hodson2022root}. This formulation ensures that relative errors are emphasized, making prediction errors for low values just as significant as those for high values. As a result, it provides a more balanced evaluation, particularly in cases where the data spans multiple orders of magnitude.

\subsection{Input Definition}

Our input data is derived from the ASHRAE Great Energy Predictor III dataset~\cite{ashrae-energy-prediction}. For each building \( b_i \) with building index $i$, the input at each time step \( t_j \) at time $j$ is defined by the feature vector:

\begin{equation}
\mathbf{x}_{t_j}^{(i)} = \begin{bmatrix} 
E_{t_j}^{(i)}, C_{t_j}^{(i)}, S_{t_j}^{(i)}, H_{t_j}^{(i)}, T_{a, t_j}^{(i)}, CC_{t_j}^{(i)}, T_{d, t_j}^{(i)}, \phi(t_j) 
\end{bmatrix}^T
\end{equation}

In this vector, \( E_{t_j}^{(i)} \), \( C_{t_j}^{(i)} \), \( S_{t_j}^{(i)} \), and \( H_{t_j}^{(i)} \) represent the meter readings for electricity, chilled water, steam, and hot water, respectively. Weather features include \( T_{a, t_j}^{(i)} \) (air temperature), \( CC_{t_j}^{(i)} \) (cloud coverage), and \( T_{d, t_j}^{(i)} \) (dew temperature). Additionally, \( \phi(t_j) \) captures timestamp-derived features such as the hour of the day, day of the week, and holiday indicators.

The comprehensive feature vector $\mathbf{x}_{t_j}^{(i)}$ encapsulates all relevant data needed for our model to capture both consumption patterns and influencing environmental factors.

\begin{figure}[htbp]
    \centering
    \includegraphics[width=\textwidth]{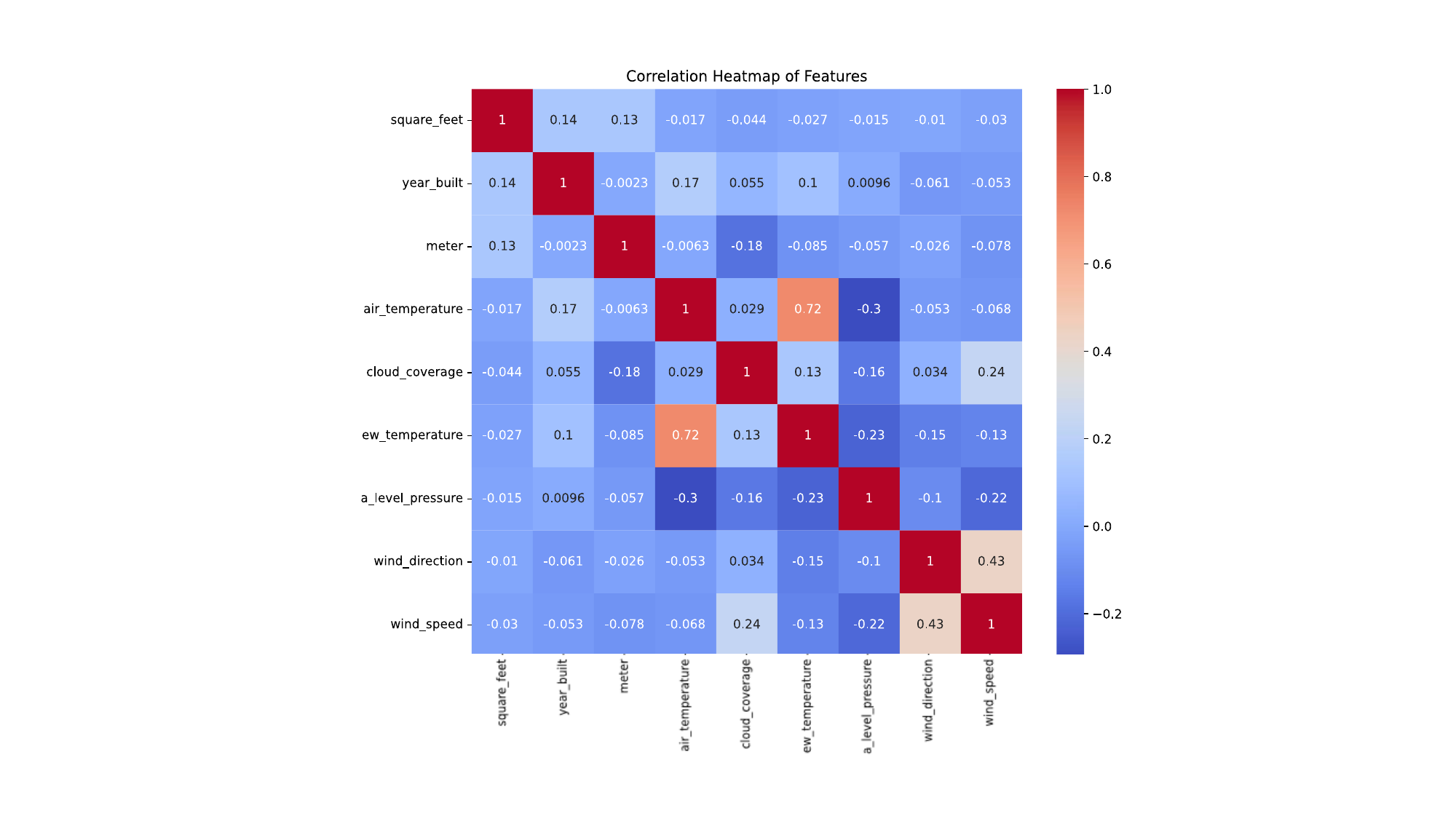}
    \Description{Heatmap of Input Correlations.}
    \caption{Heatmap of Input Correlations (same order on the x- and y-axis, thus x-axis not labeled due to repetition).}
    \label{fig:Heatmap}
\end{figure}

The heatmap depicted in Fig. 5 provides a visual representation of the correlations among the input features used for building energy consumption prediction in the ASHRAE Great Energy Predictor III dataset~\cite{ashrae-energy-prediction}. Through this heatmap~\cite{zhao2014advanced}, the strength and direction of linear relationships between pairs of variables is discerned, indicated by correlation coefficients ranging from -1 to 1. A coefficient near 1 denotes a strong positive correlation, where an increase in one variable is associated with an increase in another, whereas a coefficient near -1 suggests a strong negative correlation, indicating an inverse relationship~\cite{gu2022complex}. It uses a color gradient to represent the correlation coefficients between variables, where red indicates a strong positive correlation close to 1.0, meaning that the two variables tend to increase or decrease together. Blue represents a negative correlation close to -1.0, meaning that as one variable increases, the other tends to decrease. White or light colors indicate little to no correlation close to 0.0, suggesting no strong linear relationship between the variables. The diagonal elements are all equal to 1.0, as they represent the correlation of each feature with itself. Some notable observations from the heatmap include that air temperature and east-west temperature show a strong positive correlation of 0.72, which is expected as these two temperature-based features are closely related. Wind direction and wind speed have a moderate positive correlation of 0.43, indicating some relationship between wind movement and speed. Atmospheric pressure at sea level and air temperature show a negative correlation of -0.3, suggesting that as air temperature increases, atmospheric pressure decreases. Other features such as $square feet$ and $year built$ have weak correlations with most variables, meaning their relationships are less significant, just to explain some correlations. 

\section{S4ConvD Method}

Our proposed method builds upon the existing S4D framework~\cite{gupta2022diagonal} by introducing a novel method, S4ConvD, which reformulates the convolution operation within the state-space model to improve performance and computational efficiency (for SSMs in general see next section). While conventional CNNs~\cite{o2015introduction} or Transformers~\cite{vaswani2023attentionneed} struggle with long context windows~\cite{katharopoulos2020transformers}, SSMs retain past information efficiently through their internal state update structures~\cite{brogan1974modern, Harvey_1990}. S4D Conv combines the S4D framework~\cite{gupta2022diagonal} with convolutional mechanisms~\cite{o2015introduction} to handle long time-series data. The matrix structure of SSMs is further utilized to enable parallel and spectral processing. 
In contrast to S4D, the kernel generation is facilitated by the SSM matrix structure~\cite{gupta2022diagonal} and subsequently applied as a convolution on the input time-series signal via Fast Fourier Transform (FFT)~\cite{duhamel1990fast}.
The parameters \(\log A\) and \(A_{im}\) determine the behavior of system dynamics and influence how quickly or slowly states are updated.
By integrating with convolutions, the model focus is further directed toward relevant frequency domains.
While classical SSMs~\cite{brogan1974modern, Harvey_1990} require recursive updates, which can become numerically unstable and lead to vanishing or exploding gradients, S4~\cite{gu2021efficiently} introduced the convolutional representation of SSMs for inference. In S4D the same Cauchy Kernel is used~\cite{gupta2022diagonal}. For the proposed S4ConvD, we instead use a convolutional kernel~\cite{o2015introduction}.

\begin{table}[ht]
\centering
\small 
\caption{Comparison between S4D and S4D Conv}
\begin{tabular}{|l|p{5.0cm}|p{5.0cm}|} 
\hline
\textbf{Property} & \textbf{S4D (Structured State Space for Sequences)~\cite{gupta2022diagonal}} & \textbf{S4D Conv (Convolutional Variant)=ours} \\ \hline
\textbf{Core Principle} & Solves the recursive form of state-space equations directly in the Fourier domain for long sequences. & Performs convolution in the frequency domain, combining SSMs with CNNs for efficient processing. \\ \hline
\textbf{Computation} & Explicitly computes the state transition matrix \( A \) and output matrix \( C \) via discrete approximation. & Conducts discrete convolution over the entire sequence with FFT, similar to a CNN. \\ \hline
\textbf{Memory Requirement} & Can have high memory demands if not implemented efficiently. & Reduces memory requirements through FFT-based convolution and its possibilty to use CUDA optimization. \\ \hline
\end{tabular}
\label{tab:comparison_S4D_S4ConvD}
\end{table}

\subsection{Foundation in S4D}

\begin{wrapfigure}{R}{0.45\textwidth}
    \centering
    \includegraphics[width=0.42\textwidth]{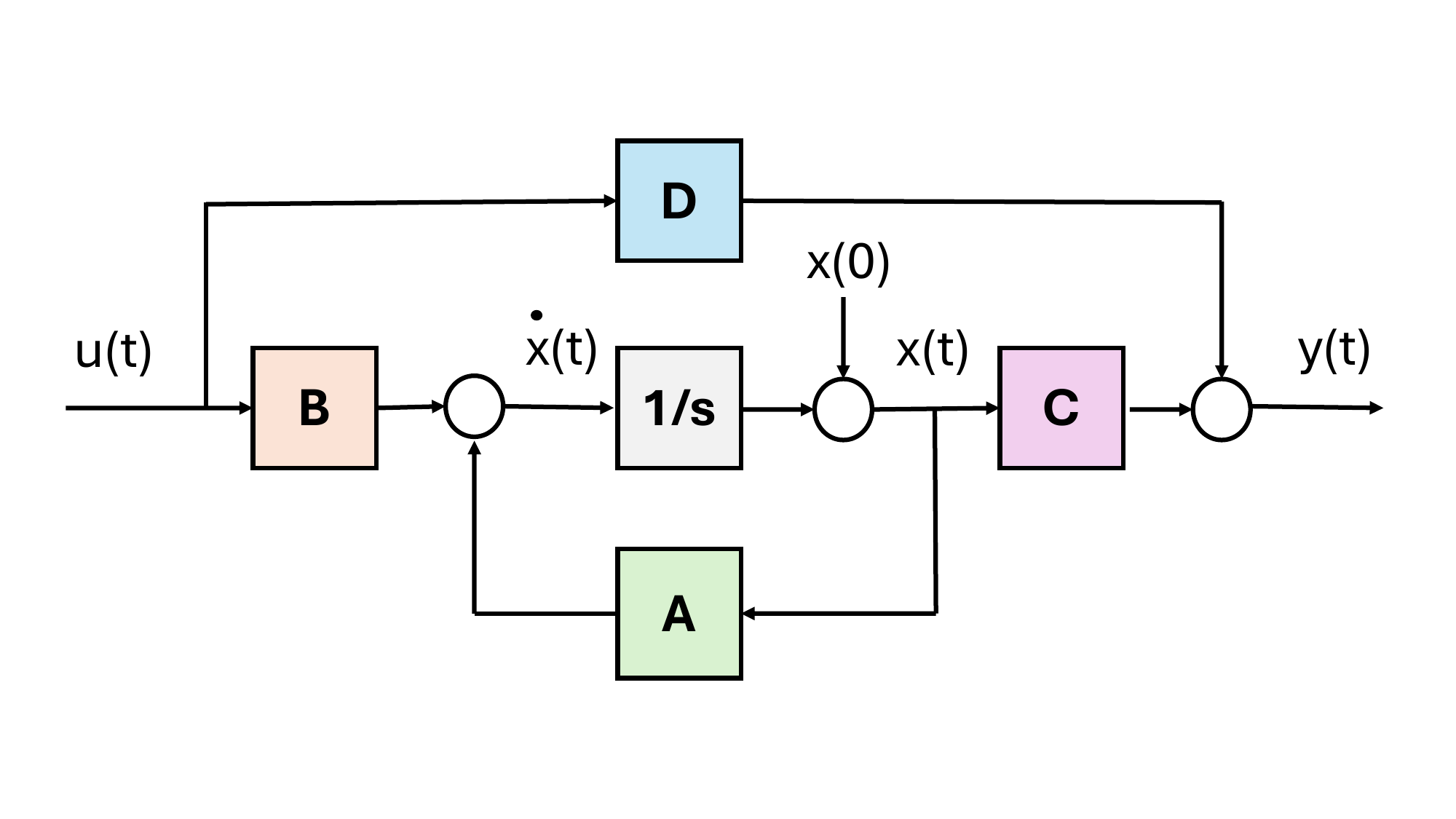} 
    \caption{Block diagram representation of SSMs.}
    \label{fig:SSM}
\end{wrapfigure}

State Space Models (SSMs) provide a systematic and mathematical framework to model the dynamics of physical systems by capturing the intrinsic relationships between input, output, and internal state variables. The block diagram representation of SSMs in Fig. 6 displays the interaction and flow of signals within the system’s structure. The primary components include input nodes $\mathbf{u}(t)$, state blocks $\mathbf{x}(t)$, output nodes $\mathbf{y}(t)$, transfer function elements represented by the matrices $\mathbf{A}$, $\mathbf{B}$, $\mathbf{C}$, and $\mathbf{D}$, as well as summation nodes, an integrator $\frac{1}{s}$, and feedback loops. 

Input nodes represent the external inputs applied to the system, which are essential for driving the system dynamics. These inputs influence the state variables, which are depicted in state blocks as central elements responsible for holding the current condition of the system, where $\mathbf{x}(0)$ represents the initial state~\cite{friedland2012control, levitin2007block, williams2007linear}. The evolution of the state is determined by the contributions of $\mathbf{B} \mathbf{u}(t)$ and $\mathbf{A} \mathbf{x}(t)$, as indicated by the left summation node in the diagram. The result of this summation is then processed through the integration block $\frac{1}{s}$, which represents the system’s memory and accumulates state changes over time.

The output is derived from the internal state through a transformation governed by $\mathbf{C} \mathbf{x}(t)$, which is computed before reaching the right summation node. This node also considers the direct contribution of $\mathbf{D} \mathbf{u}(t)$, ensuring that both state-dependent and input-dependent components contribute to the final output $\mathbf{y}(t)$. 

A key feature of the model is the feedback loop, which connects the state $\mathbf{x}(t)$ back to the summation node through the transformation by $\mathbf{A}$. This feedback mechanism highlights how the current state influences its own rate of change, a crucial aspect of capturing dynamic system behavior. By combining these elements, the block diagram provides a clear visualization of how state-space equations operate within a system, emphasizing the continuous interaction between input, state, and output variables.

State blocks are intricately linked to transfer functions, which mathematically define how system inputs transform into state changes over time (\(\dot{\mathbf{x}}(t) = \mathbf{A} \mathbf{x}(t) + \mathbf{B} \mathbf{u}(t)\)). These transformations typically involve linear operations such as multiplications by matrices, representing the system's time-invariant characteristics. Output nodes follow from the state blocks and are responsible for producing the observable outputs of the system. The output values depend not only on the state variables but may also be directly influenced by the inputs (\(\mathbf{y}(t) = \mathbf{C} \mathbf{x}(t) + \mathbf{D} \mathbf{u}(t)\)). This dual dependency is integral to the logic of SSMs, where the output is a function of the current state and the input, encapsulated by the output matrix. Feedback loops in the block diagram symbolize the control actions that adjust the input based on the output, forming a closed-loop system~\cite{friedland2012control, levitin2007block, williams2007linear}. 

The S4D model~\cite{gupta2022diagonal} is a state-space approach commonly represented by the following equations~\cite{gu2021efficiently, gupta2022diagonal}:

\begin{align}
    \mathbf{x}_{t_{j+1}}^{(i)} &= \mathbf{A} \mathbf{x}_{t_j}^{(i)} + \mathbf{B} \mathbf{u}_{t_j}^{(i)}, \\
    \mathbf{y}_{t_j}^{(i)} &= \mathbf{C} \mathbf{x}_{t_j}^{(i)},
\end{align}

where \( \mathbf{x}_{t_j}^{(i)} \in \mathbb{R}^{n_x} \) is the state vector at time step \( t_j \) for building \( b_i \), \( \mathbf{u}_{t_j}^{(i)} \in \mathbb{R}^{n_u} \) represents the input vector, and \( \mathbf{y}_{t_j}^{(i)} \in \mathbb{R}^{n_y} \) is the output vector. The matrices \( \mathbf{A} \in \mathbb{R}^{n_x \times n_x} \), \( \mathbf{B} \in \mathbb{R}^{n_x \times n_u} \), and \( \mathbf{C} \in \mathbb{R}^{n_y \times n_x} \) define the system dynamics, where \( \mathbf{A} \) governs state transitions, \( \mathbf{B} \) captures input influences, and \( \mathbf{C} \) maps states to outputs. To improve computational efficiency, \( \mathbf{A} \) is chosen to be diagonal, simplifying matrix operations and accelerating inference~\cite{gupta2022diagonal}.

\subsection{Novelty of the S4ConvD Kernel}

In the original S4 model, the computation of the convolution kernel posed significant computational challenges, especially for general state matrices $A$. The S4 model introduced a complex algorithm for Diagonal Plus Low Rank (DPLR) state matrices~\cite{gu2021efficiently}.

For diagonal state matrices $A$, however, the computation becomes numerical more stable~\cite{gupta2022diagonal}. The convolution kernel $\mathbf{K}$ is described by:

\begin{equation}
  K_\ell = \sum_{n=0}^{N-1} C_n A_n^\ell B_n \implies \mathbf{K} = (\mathbf{B} \odot \mathbf{C}) \cdot \mathbf{VL}(A)
\end{equation}

where $\odot$ denotes the Hadamard product, $\cdot$ represents matrix multiplication, and $\mathbf{VL}(A)$ is the Vandermonde matrix defined by:

\begin{equation}
 \mathbf{VL}(A)_{n,\ell} = A_n^\ell
\end{equation}

Unpacking this, $\mathbf{K}$ can be expressed using the Vandermonde matrix-vector multiplication:

\[
\mathbf{K} =
\begin{bmatrix}
B_0 C_0 & \cdots & B_{N-1} C_{N-1}
\end{bmatrix}
\begin{bmatrix}
1 & A_0 & A_0^2 & \cdots & A_0^{L-1} \\
1 & A_1 & A_1^2 & \cdots & A_1^{L-1} \\
\vdots & \vdots & \vdots & \ddots & \vdots \\
1 & A_{N-1} & A_{N-1}^2 & \cdots & A_{N-1}^{L-1}
\end{bmatrix}
\]

The approach involves materializing the Vandermonde matrix $\mathbf{VL}(A)$ and performing matrix multiplication, which requires $\mathcal{O}(NL)$ time and space. 

However, Vandermonde matrices are well-studied, and their multiplication can be theoretically computed in $\mathcal{O}(N + L)$ operations and $\mathcal{O}(N + L)$ space. Notably, Vandermonde matrices have close ties with Cauchy matrices, forming the computational core of S4’s DPLR algorithm with similar complexity characteristics.

For our novel S4ConvD method, the convolution operation is formalized differently to better integrate rapid variations in the building energy datasets. This is particularly achieved through the use of adaptive scaling and frequency adjustment that aligns with dynamic input patterns.

To effectively capture rapid variations in building energy datasets, the convolution operation in the S4ConvD method is structured to incorporate adaptive scaling and frequency adjustment. This allows the model to dynamically respond to changes in input patterns.

The convolution kernel in the S4ConvD framework is defined as:

\[
\mathbf{K}_{\text{S4ConvD}}(t) = C \times \sigma(e^{t A_{\text{adaptive}}} \cdot B_{\text{adaptive}})
\]

Here, the components are detailed as follows: \( A_{\text{adaptive}} \): A dynamically parameterized state matrix that adjusts during training to reflect real-time changes in the input data's temporal dynamics.
\( B_{\text{adaptive}} \) is an input matrix similarly parameterized to ensure that the model can incorporate variations in the input data, adapting its influence on the state transformations. 
\( \sigma(\cdot) \) is a selective non-linear transformation applied to the product \( e^{t A_{\text{adaptive}}} \cdot B_{\text{adaptive}} \). This function enhances the kernel's sensitivity to changes, differing from the constant linear scaling prevalent in traditional S4D. The non-linear transformation \( \sigma(x) \) could represent a Sigmoid function or another suitable non-linear function, such as:

\[
\sigma(x) = \frac{1}{1 + e^{-x}}
\]

This adjustment amplifies pertinent signals while suppressing noise, ensuring the kernel can dynamically reflect and adapt to new data patterns. 
Here, the $state-dim$ is 64, the $measurement-dim$ is 128, the $input-dim$ is 4, the $output-dim$ is 1, $dropout$ has been set to $0.01$, $batch-size$ has been set to 16, the $learning-rate$ of the Stochastic Gradient Descent(SGD) has been set to $0.001$, the $momentum$ for the SGD-optimizer has been set to $0.9$, the $log-interval$ has been set to 200 and the $num-epochs$ has been set to 100, equally for all models in the benchmarking. More details are available in the provided code on Github.

% \subsection{Comparison to S4D}

% Unlike the fixed convolution kernel in S4D, S4ConvD:
% \begin{enumerate}
%     \item offers a dynamic reconfiguration of state matrix elements to align with observed temporal variations, enhancing learning flexibility.
%     \item employs a streamlined computational mechanism for handling Fourier transformed sequences, further optimizing processing speeds beyond standard FFT applications used in S4D.
% \end{enumerate}

% S4ConvD introduces an adaptive scaling mechanism, enabling the model to adjust its convolutional weights in response to new data inputs. This adjustment is implemented through a learning algorithm that continuously updates the convolutional parameters based on observed changes in input sequences. S4ConvD also employs a frequency adjustment mechanism. This focuses on modifying the state-space model's response to different frequency components of the data. By engaging this mechanism, the model can strategically focus more on components that significantly contribute to energy consumption patterns. 

% TBD: Illustrative Example:
% a situation where there is an unexpected surge in energy usage due to a local event: The traditional static model might fail to capture this spike accurately, whereas S4ConvD dynamically adjusts its parameters, detecting the anomaly swiftly and adapting its predictions accordingly.

\section{Experimental Results}
In the following section, we present an analysis of the experimental results, demonstrating the capabilities and performance of our developed method. We begin with an ablation study that explores the impact of different kernel functions on the system's runtime and accuracy. Following this, we evaluate the memory footprint of our approach, highlighting its resource efficiency compared to existing methodologies.  Furthermore, we conduct a robustness validation to assess the resilience and reliability of the proposed solution. Finally, we present benchmarking results that compare our method against state-of-the-art techniques on the same dataset. Through these experimental validations, we substantiate the efficacy of our proposed method in addressing contemporary challenges in energy consumption prediction throughout different building types.

\subsection{Ablation study with different kernels}

In our ablation study, we explored the performance impact of deploying different convolutional kernels within the sequence modeling framework for energy consumption prediction. Our primary focus, the S4ConvD model, incorporates convolutional adjustments to improve sequence data interpretation. 
We compared our approach against the original S4D approach, which utilizes a Cauchy kernel renowned for its stability and capacity to smoothly model long-range dependencies. To ensure a fair comparison all hyperparameters have been set on the same value like described in section 5.2.

\begin{table}[ht]
    \centering
    \caption{Test RMSLE Comparison between S4ConvD and S4D with Cauchy Kernel}
    \begin{tabular}{lcc}
        \toprule
        \textbf{Modell} & \textbf{Test RMSLE (1\% of dataset)} \\
        \midrule
        S4D Cauchy & 4.6702 \\
        \textbf{S4ConvD (ours)} & \textbf{4.6676$\downarrow$} \\
        \bottomrule
    \end{tabular}
\end{table}

\subsection{Memory Evaluation}

% \begin{wrapfigure}{L}{0.5\textwidth}
%     \centering
%     \includegraphics[width=0.48\textwidth, trim=0 280 0 280, clip]{samples/b4a30a0c-9c38-46df-a614-69c8ac0b526d.pdf} 
%     \caption{Process Memory Available (MB) in comparison between S4D with Cauchy Kernel and S4ConvD with Time in minutes on the x-axis and Process Memory Available in MB on the y-axis.}
%     \label{fig:Processing}
% \end{wrapfigure}

% \begin{wrapfigure}{R}{0.5\textwidth}
%     \centering
%     \includegraphics[width=0.48\textwidth, trim=0 280 0 280, clip]{samples/792cf953-c32b-4bd0-bba8-9367a4c022ad.pdf} 
%     \caption{Disk Utilization (GB) in comparison between S4D with Cauchy Kernel and S4ConvD with time in minutes on the x-axis and Disk Utilization in GB on the y-axis.}
%     \label{fig:Disk_Util}
% \end{wrapfigure}

In our comparative memory analysis between S4D and S4ConvD over 5 epochs in Fig. 7 (a), we identified a reduction in process memory usage when utilizing the S4ConvD method. This efficiency was consistently observed during both the training and inference phases. By integrating memory tracking capabilities through Weights and Biases~\cite{jocher2021ultralytics}, we were able to monitor memory allocation and usage dynamically. The streamlined architecture of S4ConvD, which leverages FFT-based convolutions, contributes to this reduction by minimizing the storage demands typically associated with state-space computations. As can be observed from the figure, the process memory available in MB is lower for S4ConvD, which means that the RAM usage is lower. The observed pattern, where process memory usage increases at the start of each epoch and gradually decreases towards the end, can be explained by the dynamic nature of memory allocation during model training. At the beginning of each epoch, the computational demands increase as the model loads new batches of data and associated parameters into memory. This initialization phase requires substantial memory resources to accommodate the data structures involved in processing new batches, gradient accumulation, and other temporary computations. As the epoch progresses, memory optimization mechanisms such as memory recycling and garbage collection come into play, freeing up resources by deallocating memory associated with completed computations and intermediate variables no longer in use. This resource management strategy leads to a gradual decline in memory usage as the epoch advances. 

\begin{figure}[ht]
    \centering
    \begin{subfigure}{0.48\textwidth}
        \centering
        \includegraphics[width=\textwidth, trim=0 280 0 280, clip]{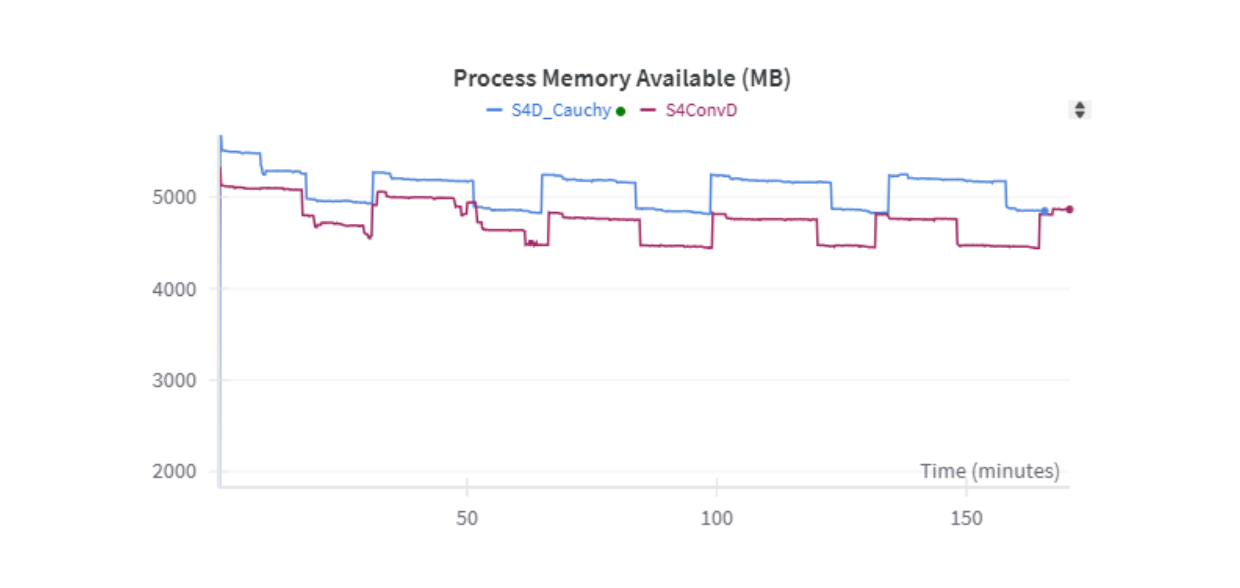}
        \caption{Process Memory Available (MB) for S4D with Cauchy Kernel and S4ConvD. Time in minutes on the x-axis and Process Memory Available in MB on the y-axis.}
        \label{fig:Processing}
    \end{subfigure}
    \hfill
    \begin{subfigure}{0.48\textwidth}
        \centering
        \includegraphics[width=\textwidth, trim=0 280 0 280, clip]{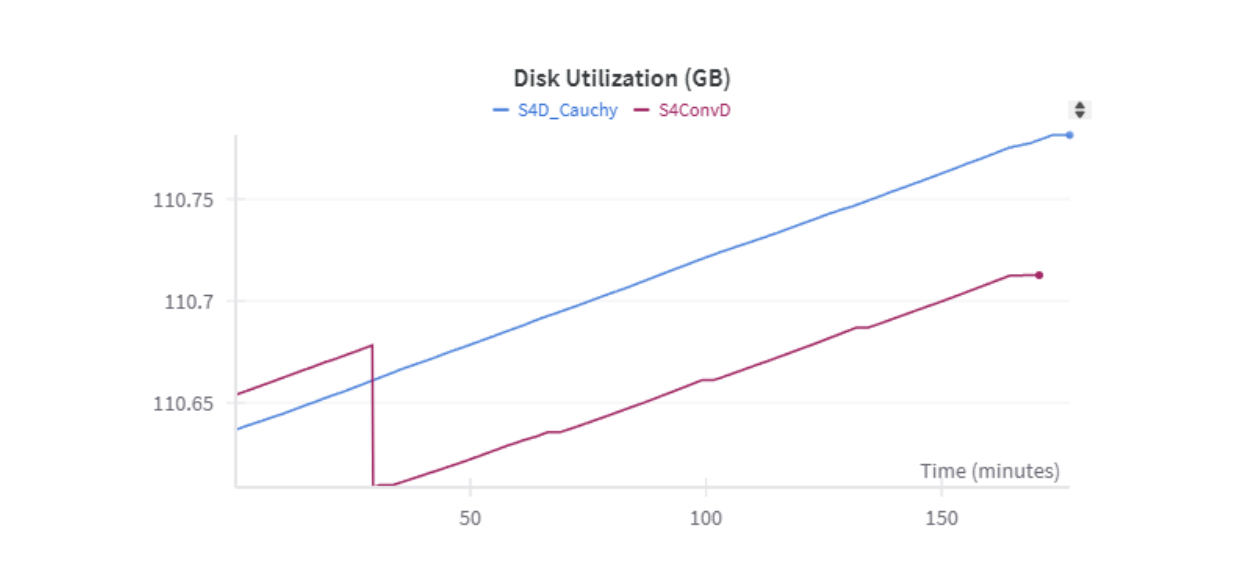}
        \caption{Disk Utilization (GB) for S4D with Cauchy Kernel and S4ConvD. Time in minutes on the x-axis and Disk Utilization in GB on the y-axis.}
        \label{fig:Disk_Util}
    \end{subfigure}
    \caption{Comparison of Process Memory and Disk Utilization between S4D and S4ConvD.}
\end{figure}

Continuing from the analysis of memory usage, our evaluation extends to the comparison of disk utilization between the S4D and S4ConvD models as illustrated in Fig. 7 (b). The S4ConvD model demonstrates an improvement in disk utilization efficiency. This is largely attributable to its FFT-based convolutional strategy, which minimizes the frequency and volume of disk I/O operations. During both training and inference, the reduced disk access requirements of S4ConvD lessen the model's dependency on persistent storage, thereby decreasing the disk utilization. Lower disk utilization also indicates that S4ConvD processes data more efficiently in-memory, relying less on disk storage for intermediary computational tasks. Consequently, this translates to a decrease in latency that can arise from disk read/write operations, providing a smoother and faster execution flow. The observed temporary drop to zero in disk utilization at the 30-minute mark for the S4ConvD model can be attributed to its reliance on in-memory computations and the high-speed processing capabilities inherently tied to Fast Fourier Transforms (FFT). These architectures are designed to maximize memory utilization, thereby minimizing dependency on slower disk operations. By prioritizing in-memory computations, the model reduces latency and enhances processing efficiency, which may lead to temporary pauses in disk I/O operations during intensive data processing phases. As a result, the system can manage data more effectively without frequent disk access, explaining the brief drop during this period.

\subsection{Robustness Validation}
In order to ensure the robustness of the results and validate the consistency of our findings, we conducted repetitive experiments with both the S4D Cauchy Kernel and the S4ConvD approach. Each model was trained and evaluated on subsets of the dataset, specifically using 1\% and 5\% of the data. These experiments were repeated ten times for each kernel. The results demonstrated a remarkable consistency, as no significant deviations were observed in the performance metrics for either kernel across the repeated runs (e.g. the RMSLE remained 4.6676 for S4ConvD for all 10 runs with 1\% of the data, see Table 2). This demonstrates, that the observed performance is stable and not an artifact of random sampling variability. 

\subsection{Benchmarking Results}

Based on the defined data split, we re-implemented the top three submissions and some of the best other used models from the challenge on the ASHRAE Great Energy Predictor III competition leaderboard to serve as a benchmark against our proposed S4ConvD.

\begin{table}[ht]
\centering
\caption{Top submissions in the ASHRAE Great Energy Predictor III Competition}
\begin{tabular}{|l|c|}
\hline
 \textbf{Model} & \textbf{RMSLE} \\ \hline
\textbf{S4DConv (ours)} & \textbf{0.395} \\ \hline
S4D & 0.425 \\ \hline
Decision Tree & 0.607 \\ \hline
Ensemble LGB-CatBoost-XGBoost & 1.232 \\ \hline
Light GBM & 1.292 \\ \hline
Linear Regression & 1.381 \\ \hline
Lasso Estimator & 1.381 \\ \hline
SGD Regression & 1.381 \\ \hline
Ridge & 1.384 \\ \hline
Elastic Net & 1.473 \\ \hline
SVR & 2.298 \\ \hline
\end{tabular}
\end{table}

As Table 3 shows, our proposed S4ConvD method uses an adaptive scaling and frequency adjustment within its convolution kernel. This allows it to respond dynamically to variations in input data, hence capturing real-time changes in energy consumption patterns more effectively than static methods.
Although our method did not incorporate extensive data preprocessing techniques, such as robust feature engineering or anomalous data filtering, it nonetheless achieved superior performance in comparison to the ASHRAE Great Energy Predictor III competition submissions. By employing an adaptive convolution kernel that dynamically adjusts to rapidly changing input sequences, our approach inherently captures the essential patterns and dependencies within the dataset without requiring labor-intensive preprocessing steps. It also shows to be slightly better than the original S4D approach with its Cauchy kernel.
The Decision Tree model had an RMSLE of 0.607, which was better than the winning submissions in the challenge. The ensemble model combining LightGBM, CatBoost, and XGBoost, despite incorporating a variety of predictors performed notably lower than our S4ConvD, with an RMSLE of 1.232. LightGBM and other linear regression-based approaches like Lasso, Ridge, and Elastic Net exhibited limitations in inherently capturing nonlinear dependencies even with feature engineering. Their RMSLE scores, ranging from 1.292 to 1.473, reach middle scores.
On the extreme, SVR’s high RMSLE of 2.298 indicates challenges in scalability and sensitivity to variations in the data set that the S4ConvD efficiently managed. 

\section{CUDA Optimization}

In CUDA optimization, several key techniques are employed to enhance performance by improving memory access patterns and computational efficiency~\cite{fauzia2015characterizing}. Block Tiling~\cite{van2011optimizing, bastem2019tiling} further refines efficiency by organizing data into chunks, or tiles, that are processed by thread blocks. Such a scheme optimizes data parallelism and minimizes cache misses. In order to achieve a high usage, we set the batch-size to 8192 for all experiments on this GPU. For our experiments, we used a NVIDIA Tesla P100 GPU, with the specifications displayed in Table 4. 

\begin{table}[htbp]
\centering
\caption{Specifications of the NVIDIA Tesla P100 GPU}
\begin{tabular}{|l|c|}
\hline
\textbf{Metric} & \textbf{Value} \\ \hline
Name & NVIDIA Tesla P100 \\ \hline
Compute Capability & 6.0 \\ \hline
Max Threads per Block & 1024 \\ \hline
Max Threads per Multiprocessor & 2048 \\ \hline
Threads per Warp & 32 \\ \hline
Warp Allocation Granularity & 4 \\ \hline
Max Registers per Block & 65536 \\ \hline
Max Registers per Multiprocessor & 65536 \\ \hline
Register Allocation Unit Size & 64 \\ \hline
Register Allocation Granularity & Warp \\ \hline
Total Global Memory & 16280 MB \\ \hline
Max Shared Memory per Block & 48 KB \\ \hline
CUDA Runtime Shared Memory Overhead per Block & 512 B \\ \hline
Shared Memory per Multiprocessor & 65536 B \\ \hline
Multiprocessor Count & 56 \\ \hline
Max Warps per Multiprocessor & 64 \\ \hline
\end{tabular}
\label{tab:tesla_p100_specs}
\end{table}

In CUDA, a warp consists of 32 threads, so the tile size should be a multiple of 32 to ensure full warp utilization. Each block can have up to 1024 threads, aligning with potential block configurations for efficiency. The maximum shared memory per block is 48 KB, which limits the tile configuration to not exceed this capacity. For a block using 8192 bytes of shared memory and 512 bytes for runtime overhead, the total is 8704 bytes per block, allowing up to 7 blocks in shared memory.

With 1024 threads per block and a max of 2048 threads per multiprocessor, two blocks can fit in shared memory. Each thread uses 37 registers, totaling 38848 registers per block. Given 65536 registers per multiprocessor, this is the limiting factor as only one block fits per shared memory. This results in approximately 50\% occupancy, with 32 out of 64 possible active warps per multiprocessor. Consequently, a tile size of 32 is set for Block Tiling on the NVIDIA Tesla P100 GPU to optimize utilization.

\begin{table}[htbp]
\centering
\caption{Performance Impact of Various CUDA Optimizations on S4ConvD method}
\begin{tabular}{|l|c|c|c|c|}
\hline
\textbf{Used Technique}  & \textbf{Time/epoch (s)} & \textbf{Timered.(\%)} & \textbf{GPU Memory(\%)} & \textbf{Usage Improv.(\%)} \\ \hline
Naive Kernel  &  01:05 &   0 &  9.8 GiB (63\%) &  0 \\ \hline
Block Tiling (32) &  00:42$\downarrow$ &  35.38$\uparrow$  &  9,8 GiB (99\%) & 36$\uparrow$ \\ \hline
\end{tabular}
\label{tab:cuda_optimizations}
\end{table}

This CUDA optimization process exemplifies the importance of algorithmic structure in deriving benefits from the GPU's architecture. Using S4ConvD allowed an easier transition and application of CUDA optimization techniques. In contrast, a Cauchy Kernels, due to their matrix structure and specialized computational needs, typically cause scattered memory accesses. This scattering complicates efforts to implement efficient tiling strategies. In Table 5 we show, that optimizing the kernel for higher arithmetic intensity (fewer memory operations, more computation) and ensuring efficient shared memory usage can maximize performance within these resource constraints about 36 \%. 

% \section{Alignment with emerging standards for energy control}

% The adoption of the S4ConvD methodology in predicting building energy consumption aligns well with emerging standards for energy control~\cite{janda1994worldwide}, which increasingly emphasize the integration of flexible, adaptive technologies that can respond in real-time~\cite{yin2016joint} to dynamic environmental conditions~\cite{ingelrest2010sensorscope}. As urban infrastructures grow more interconnected and complex, there is a growing need for energy management solutions~\cite{erickson2014occupancy} that are not only accurate but also capable of quick adaptation to varying conditions~\cite{brodin2007theoretical}. The S4ConvD approach inherently supports these requirements by offering real-time adaptability and precision in modeling energy patterns. Moreover, as energy control standards~\cite{janda1994worldwide} evolve to favor sustainability and efficiency, methods like S4ConvD, which streamline computational resources and optimize predictions without a heavy reliance on preprocessing or manual data intervention, lead the way in cost-effective and scalable solutions. By predicting consumption patterns and potential savings, our method could be used to meet stricter energy efficiency goals, supporting the development of smart grids~\cite{ahmed2012implementation}, and facilitating the integration of renewable energy sources~\cite{blacha2019dynamic}. 

\section{Conclusion}
In this paper, we introduced S4ConvD, an convolutional variant of Deep State Space Models, to address the challenges inherent in predicting energy consumption in smart buildings. By utilizing adaptive scaling and frequency adjustments, S4ConvD captures changing patterns in the multivariate time-series data of the smart meter sensor network. Our evaluations on the ASHRAE Great Energy Predictor III dataset demonstrate that S4ConvD surpasses existing benchmarks in predictive accuracy, showcasing its capability to generalize across diverse urban energy infrastructures.

The efficiency of S4ConvD is highlighted by its reduced memory and disk utilization, making it an option for deployment in resource-constrained real-world environments. These efficiencies are primarily due to the use of FFT-based convolutions, which streamlined computational processes without sacrificing model performance in our task. 

Furthermore, the integration of CUDA optimization techniques, such as Global Memory Coalescing and Block Tiling, further amplifies the performance of S4ConvD. By leveraging the architectural strengths of the NVIDIA Tesla P100 GPU, we achieved a significant reduction of 35.38\% in computational time per epoch, alongside maximal memory utilization at 99\%.  
Its design principles promote resource-efficient model execution, enhancing both energy forecasting and the potential integration of renewable energy sources into smart grid systems.

By minimizing reliance on extensive preprocessing and emphasizing dynamic adaptability, S4ConvD presents a scalable solution for energy-efficient sensor networks. In future work we will explore the integration of S4ConvD into broader smart city frameworks, further validating its impact on energy management systems. Our work also demonstrates significant synergies between machine learning and energy modeling, particularly in enhancing prediction accuracy and response times in smart grids. In future research we will also focus on the integration of S4ConvD in real-time with IoT platforms, enabling seamless data flow and analysis for instant decision-making in urban infrastructures.

\begin{acks}
We thank the German Federal Ministry of Education and Research (BMBF) under the grant number 13GW0586F for funding research about state space models for time-series forecasting.
\end{acks}

\bibliographystyle{ACM-Reference-Format}
\bibliography{sample-base}

\appendix

\end{document}